\documentclass{esannV2}
\usepackage[dvips]{graphicx}
\usepackage[latin1]{inputenc}
\usepackage{amssymb,amsmath,array}
\usepackage{subfig}
\usepackage{bm}

\usepackage{algpseudocode} 
\usepackage{algorithm}

\usepackage{textpos}

\newcommand{\x}{\mathbf{x}}
\newcommand{\z}{\mathbf{z}}
\renewcommand{\u}{\mathbf{u}}
\newcommand{\s}{\mathbf{s}}
\newcommand{\y}{\mathbf{y}}
\newcommand{\W}{\mathbf{W}}
\newcommand{\Win}{\W_{in}}
\newcommand{\Wout}{\W_{out}}
\newcommand{\Input}{\mathcal{U}}
\renewcommand{\S}{\mathcal{X}}
\newcommand{\Output}{\mathcal{Y}}
\newcommand{\Fh}{\hat{F}}
\newcommand{\R}{\mathbb{R}}

\begin{document}
\title{Chasing the Echo State Property}

\author{Claudio Gallicchio
\vspace{.3cm}\\
Department of Computer Science, University of Pisa\\
Largo B. Pontecorvo, 3 - 56127 Pisa, Italy
}
\maketitle

\begin{textblock*}{150mm}(-2cm,-6cm)
\noindent
\emph{This is a preprint version of the following paper, published in the proceedings of ESANN 2019:}\\
Claudio Gallicchio (2019) Chasing the Echo State Property. In: ESANN 2019 proceedings, European Symposium on Artificial Neural Networks, Computational Intelligence and Machine Learning. Bruges (Belgium), 24-26 April 2019, i6doc.com publ., ISBN 978-287-587-065-0.
\end{textblock*}

\begin{abstract}
Reservoir Computing (RC) provides an efficient way for designing dynamical recurrent neural models.
While training is restricted to a simple output component, the recurrent connections are left untrained after
initialization, subject to stability constraints specified by the Echo State Property (ESP).
Literature conditions for the ESP typically fail to properly account for 
the effects of driving input signals, often limiting the potentialities of the RC approach.
In this paper, we study the fundamental aspect of asymptotic stability of RC models in presence of driving input, 
introducing an empirical ESP index that enables to easily analyze the stability regimes of reservoirs. 
Results on two benchmark datasets reveal interesting insights on the dynamical properties of input-driven reservoirs,
suggesting that the actual domain of ESP validity is much wider than what  
covered by literature conditions commonly used in RC practice.
\end{abstract}

\section{Introduction}
Reservoir Computing (RC) \cite{Lukosevicius2009} denotes a class of Recurrent Neural Networks (RNNs) featured by untrained recurrent connections. Essentially, RC networks are composed by a non-linear dynamical system, called \emph{reservoir}, and a (typically) linear output layer, called \emph{readout}. Training is restricted to just the readout component, 
leading to an extremely efficient way of designing RNNs. When the dynamical reservoir is described in terms of discrete-time iterated maps, the model is known under the name of Echo State Network (ESN) \cite{Jaeger2004}. Besides the striking efficiency advantage, the study of ESNs allows investigating the architectural properties of RNN systems apart from (or prior to) learning of recurrent connections. A major question arising in this context is how to enforce stability of dynamics at the stage of network initialization. This is typically approached by requiring the reservoir to satisfy an asymptotic stability property called Echo State Property (ESP) \cite{Jaeger2001}.
Known literature conditions for the ESP mainly focus on analyzing algebraic properties of the recurrent weight matrix, but typically fail to effectively take into account also the fundamental effects of driving input signals on the reservoir behavior. 
In this paper, we take a empirical perspective to the study of stability of input-driven dynamical reservoir systems and propose a simple approach to assess the validity of the ESP for given input signals. The proposed methodology is demonstrated on two benchmark real-world datasets, showing interesting insights on the dynamical properties of ESN models.

\section{Echo State Property Index}
We study the behavior of non-linear discrete-time dynamical reservoirs, described in terms of a state-space model by a state transition function\footnote{Bias terms are not reported for conciseness.} $F$:
\begin{equation}
\label{eq.reservoir}
\begin{array}{l}
F:\S \times \Input \to \S \\
\x(t) = F(\x(t-1),\u(t)) \\
\quad\quad =f(\W \x(t-1) + \Win \u(t)),
\end{array}
\end{equation}
where $\S \subset \R^{N_R}$ is the state space, $\Input \subset \R^{N_U}$ is the input space, $\x(t) \in \S$ is the  state at time $t$, and $\u(t) \in \Input$ is the external input at time $t$. The reservoir parameters are collected in
$\W \in \R^{N_R \times N_R}$ and $\Win \in \R^{N_R \times N_U}$, respectively denoting the recurrent reservoir matrix and the input-to-reservoir weight matrix. Crucially, both $\W$ and $\Win$ are kept untrained after initialization.
In our notation, $N_R$ and $N_U$ are used to indicate the reservoir size and the input dimension, respectively.
We use $f$ to represent the element-wise application of a non-linear activation function. In our analysis, we shall assume that both the input and the state spaces are compact sets, where the latter condition is always ensured when $f$ is a bounded non-linearity. In particular, as common is ESN settings, we use $f \equiv \tanh$.
The reservoir system is coupled with a readout tool, which at every time $t$ computes the output of the ESN, i.e. $\y(t) \in \Output$ $\subset \R^{N_Y}$, through the linear combination: $\y(t) = \Wout \x(t)$. We use $\Output$ to denote the output space. The elements in $\Wout$ represent the only trainable weights of the model, which are typically found in closed form by using direct methods, e.g. ridge-regression.

Given an input signal $\s = [\u(1), \u(2), \ldots] \in \Input^*$ and an initial state $\x(0) = \x_0 \in \S$, the dynamics of the reservoir ruled by eq.~\ref{eq.reservoir} evolve describing an orbit $O(\x_0) = \{\x(0), \x(1), \ldots\}$. Typically, a zero initial state is used, i.e. $\x_0 = \mathbf{0}$. In this context, it is also convenient to introduce an iterated version of eq.~\ref{eq.reservoir}, i.e. $\Fh: \S \times \Input^{*} \to \S$, such that given any initial state $\x_0$ and any input sequence $\s$, $\Fh(\x_0, \s)$ is the network state that results after iterated application of eq.~\ref{eq.reservoir} under the influence of the input signal $\s$. 
As the reservoir's parameters in $\W$ and $\Win$ do not undergo a training process, it is fundamental to impose constraints on their initialization, in order to ensure dynamical stability in applications. This requirement is typically expressed by means of the Echo State Property (ESP) \cite{Jaeger2001}.
An ESN whose reservoir dynamics are governed by eq.~\ref{eq.reservoir} is said to satisfy the ESP whenever for every initial conditions 
$\x_0, \z_0 \in \S$, and for any input sequence of length $N$, i.e. $\s_N = [\u(1), \ldots, \u(N)]$, it holds that:
\begin{equation}
\label{eq.esp}
\| \Fh(\x_0,\s_N) - \Fh(\z_0,\s_N) \| \to 0 \textrm{ as } N \to \infty.
\end{equation}
Essentially, the ESP expresses the requirement of reservoir dynamics characterized by global asymptotic (Lyapunov) stability under the influence of the driving input. In other words, the orbit of the reservoir in the state space should be asymptotically determined uniquely by the input signal, and the influence of initial conditions should progressively fade away.

Literature works provided two kind of conditions for the ESP, typically in relation to algebraic properties of $\W$.
Sufficient conditions for the ESP were initially developed by studying eq.~\ref{eq.reservoir} as a contraction mapping. In such a case, indeed, the reservoir can be shown to satisfy the ESP for any input sequence \cite{	Gallicchio2011NN}. By nature, this kind of conditions is very restrictive, and the tighter bound known to date 
 is related to the diagonal Schur stability of $\W$ \cite{Yildiz2012,Buehner2006}. More recently, the work in \cite{Manjunath2013} shifted the attention to analyzing the properties of the reservoir dynamics under the influence of 
driving external signals, providing an improved formulation of the sufficient condition for the ESP linked to an input. These contributions are summarized in the \emph{sufficient condition for the ESP} that we give in the following eq.~\ref{eq.sufficient}:
\begin{equation}
\label{eq.sufficient}
\begin{array}{l}
\bullet\; \W \textrm{ is diagonally Schur stable, } or\\
\bullet\; \lim_{t\to\infty} \sup \frac{1}{t} \sum_{i = 1}^{t} (C(t) - (1+\ln(2)))\, I\{C(t) \geq 2\} > \frac{\ln(\|\W\|)}{2},
\end{array}
\end{equation}
where $C(t) = \min(|\Win \u(t)|)$, and $I$ is the indicator function that is $1$ when its argument is true, and $0$ otherwise. 
Besides, a well known necessary condition for the ESP \cite{Jaeger2001}
originates from the study of asymptotic stability of the reservoir system in eq~\ref{eq.reservoir}, linearized around the zero state and in presence of zero input.  Under these assumptions, a \emph{necessary condition for the ESP} is given in the following eq.~\ref{eq.necessary}:
\begin{equation}
\label{eq.necessary}
\rho(\W) < 1,
\end{equation}
where $\rho(\cdot)$ denotes the \emph{spectral radius} of its matrix argument, i.e. the maximum among the absolute values of its eigenvalues. Although the condition in eq.~\ref{eq.necessary} is often used in most practical cases, it is worth noticing that for non-trivial (i.e., non-zero) input signals it is neither sufficient nor necessary for the ESP. 

In this paper, we take a concrete perspective to the problem of assessing if the ESP is satisfied for a specific input signal by a given dynamical reservoir, aiming at estimating its degree of global asymptotic stability. To this end, we introduce an \emph{ESP index}, which expresses the average deviation of reservoir orbits generated from random initial conditions to a reference orbit achieved starting from the zero state. Essentially, if there exists a unique orbit that globally attracts all the reservoir dynamics under the influence of the same input signal (i.e., if the ESP index is zero), then we can claim that the ESP is empirically satisfied. 
The steps used to compute the proposed ESP index are detailed in Algorithm~\ref{alg.index}. 
In evaluating both the randomly initialized and the reference reservoir orbits, we consider dynamics ruled by eq.~\ref{eq.reservoir} driven by the same $L$ time-steps long 
input signal, discarding the states in the first $T$ time-steps as initial transient. The ESP index is averaged over $P$ randomly generated initial conditions, and deviation between orbits is measured in terms of Euclidean distance.
\begin{algorithm}[h!] 
\caption{ESP index} 
\label{alg.index} 
\begin{algorithmic}
\Function{ESPindex}{P,T, $[\u(1),\ldots,\u(L)]$}
\State Initialize the reservoir to $\x_0 = \mathbf{0}$
\State Iterate eq.~\ref{eq.reservoir} with $\u(1), \ldots \u(L)$ and compute $O(\x_0)$
\For{$i:=1 \to P$}
\State Initialize the reservoir to a random initial state $\z_0 \in \S$
\State Iterate eq.~\ref{eq.reservoir} with $\u(1), \ldots \u(L)$ and compute $O(\z_0)$
\For{$t:=T+1 \to L$}
\State $\delta_i(t-T) = \| \Fh(\x_0, [\u(1),\ldots,\u(t)]) - \Fh(\z_0, [\u(1),\ldots,\u(t)]) \|$
\EndFor
\State $\Delta_i = \frac{1}{L-T} \sum_{j = 1}^{L-T} \delta_i(j)$
\EndFor
\State \Return $\frac{1}{P} \sum_{j = 1}^{P} \Delta_j$
\EndFunction
\end{algorithmic}
\end{algorithm}

\section{Experiments}
\begin{figure}[h!]
	\centering
		\includegraphics[width=0.9\textwidth]{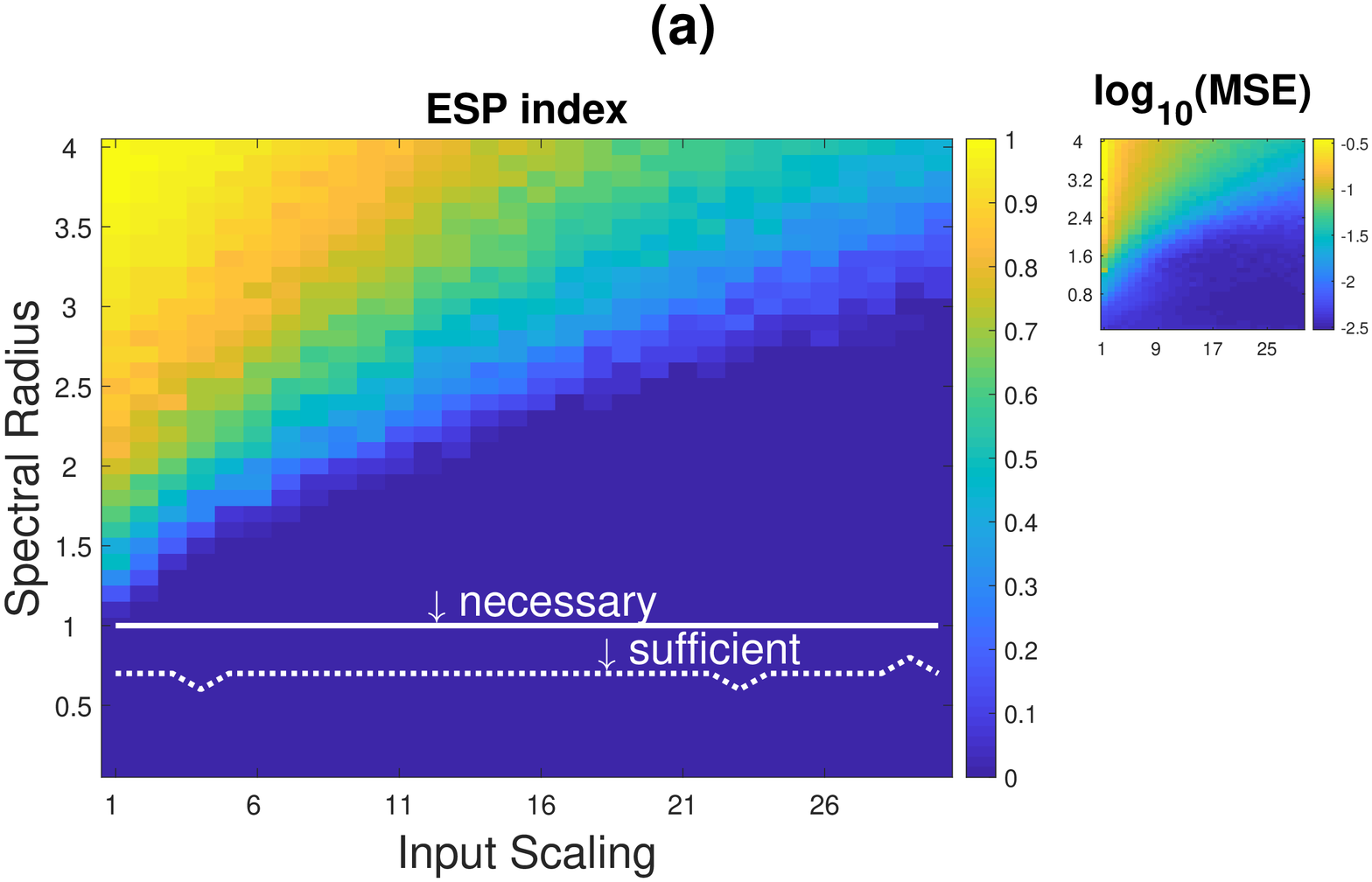}\\
		\includegraphics[width=0.9\textwidth]{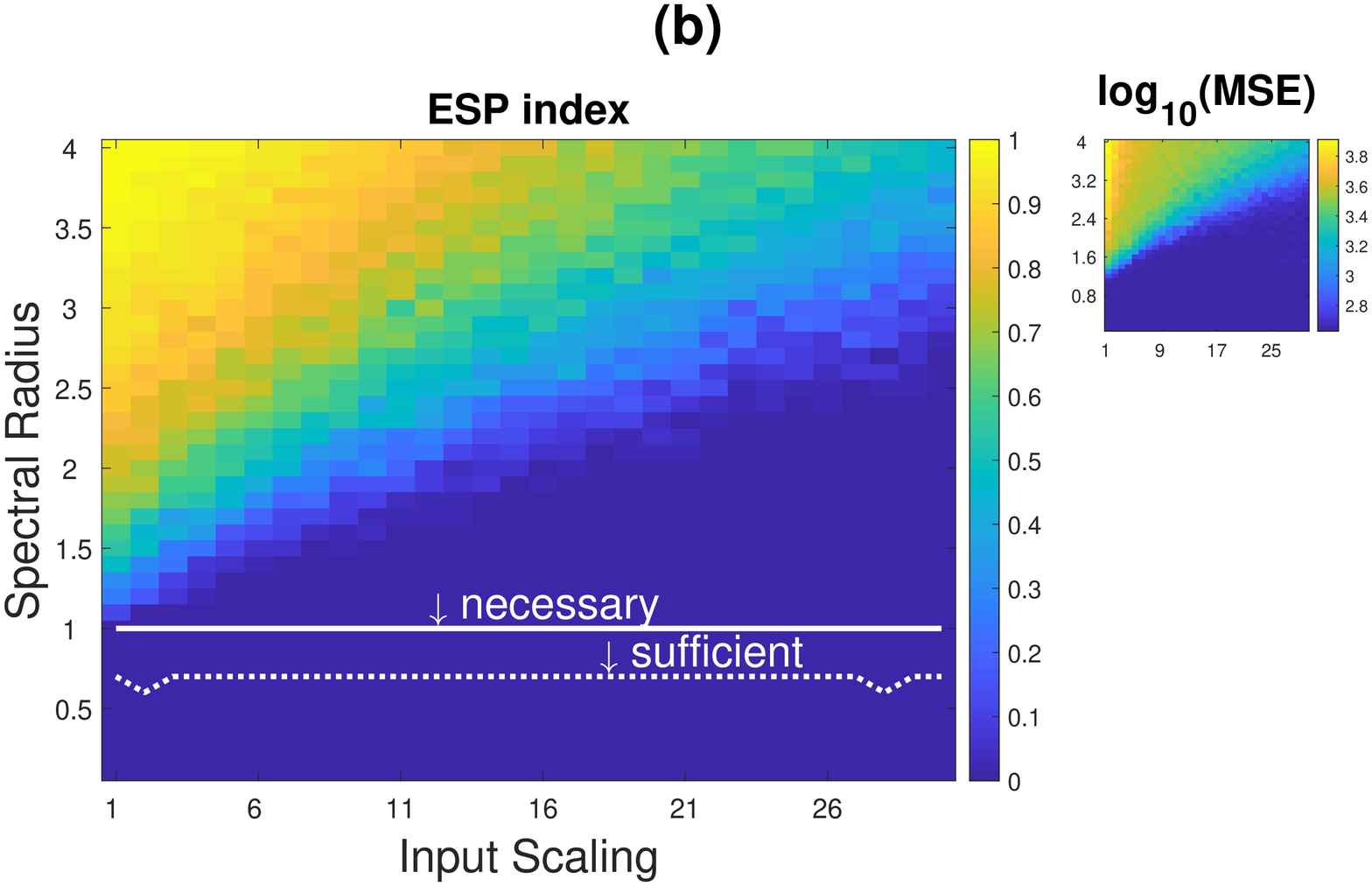}
	\caption{ESP index (main plot) and corresponding test error (smaller plot) on Laser $\mathbf{(a)}$
	and Sunspot $\mathbf{(b)}$ datasets.}
	\label{fig.results}
\end{figure}

We applied the introduced methodology for empirical assessment of the ESP considering two real-world benchmark datasets in the RC area. The first one is the Santa Fe \emph{Laser} dataset, which consists in a time-series of normalized sampled intensities from an infra-red laser in chaotic regime. The second one is a \emph{Sunspot} dataset, i.e. a time-series of monthly averaged sunspot numbers\footnote{Taken from http://sidc.be/silso/home.}, considered from January 1749 to September 2018.
In our experimental setting, we considered ESNs with 100 (fully connected) reservoir units. We initialized reservoir matrices in eq.~\ref{eq.reservoir} randomly from a uniform distribution over $[-1,1]$, re-scaling them to control two major reservoir hyper-parameters, namely the \emph{spectral radius} $\rho(\W)$, and the \emph{input scaling}  $\|\Win\|$. 
We explored values of $\rho(\W)$ between $0.1$ and $4$, with step $0.1$, and values of $\|\Win\|$ from $1$ to $30$, with step $1$.
For all the considered reservoir configurations, we computed the value of the ESP index according to Algorithm~\ref{alg.index}, using the first $L = 1000$ time-steps of the input from each dataset, 
with a transient of $T = 500$, and a number of random initial states of $P = 50$. For every configuration we also assessed whether the corresponding reservoir satisfies the sufficient and the necessary conditions, respectively expressed by eq.~\ref{eq.sufficient} and \ref{eq.necessary}.
Moreover, in the explored settings, we  evaluated the performance of the ESNs in the next-step prediction tasks defined on the considered datasets. For the Laser task we used the first 5000 time-steps for training and the remaining 5092 for test. For the Sunspot task, training and test sets contained respectively 3000 and 237 time-steps, and input values were scaled by a factor of 1000. Readout training was performed by ridge-regression, with regularization coefficient determined using the OCReP method \cite{Cancelliere2015}.
As common in ESNs applications, training discarded an initial washout (of length 1000 in our experiments).
For each reservoir configuration, all results were averaged over a number of $20$ realizations (for random initialization).

Fig.~\ref{fig.results} shows the achieved results on the Laser (Fig.~\ref{fig.results}$\mathbf{(a)}$) and Sunspot (Fig.~\ref{fig.results}$\mathbf{(b)}$) datasets. The main plot shows the averaged ESP index values achieved in correspondence of the explored reservoir settings (scaled to a maximum value of $1$ for graphical representation convenience), where darker colors indicate smaller values, and a value of $0$ indicates that the ESP is empirically verified. For convenience, upper bounds to the set of configurations satisfying the necessary and the sufficient conditions are graphically shown as well. The smaller plot shows the test errors achieved on the prediction task in 
correspondence of the same reservoir configurations, expressed in terms of $log_{10}(\textrm{MSE})$.

Results in Fig.~\ref{fig.results} are illuminating in several ways.
First of all, we can observe that in both the analyzed cases 
the set of  hyper-parametrizations that empirically satisfy the ESP
comprises reservoir configurations well beyond those encompassed by the commonly adopted ESP conditions. 
Interestingly, the ESP index plots also empirically confirm  the intrinsic stabilizing effect of increasing input sizes on RNN dynamical regimes. Extending our analysis also to the performance plots we see that the domain of empirical ESP validity, bounded by a region characterized by a sharp stable-unstable transition of recurrent dynamics, nicely correspond to the set of reservoir configurations with lower generalization error (darker points in the smaller plots), with a sharp performance decay around the ESP validity border.
In this sense, our analysis also reveals that a large portion of ``good'' reservoirs are usually neglected in common RC practice.

\section{Conclusions}
We have studied global asymptotic stability constraints for untrained reservoir dynamics driven by external input.
Assuming a concrete perspective in our analysis, we have proposed an empirical ESP index, which enables to easily assess whether an ESN shows stable dynamics in presence of given input signals.
The proposed approach has been demonstrated on RC benchmark datasets revealing interesting insights on the actual stability properties of input-driven reservoirs. Looking ahead, the work described in this paper allows us to foresee automatic task-dependent approaches for effectively setting RC configurations.
Besides, it would be a useful starting point to  drive theoretically-oriented studies aiming at unleashing the true potentialities of RC models in real-world applications.

\begin{footnotesize}

\bibliographystyle{unsrt}
\bibliography{references}

\end{footnotesize}

\end{document}